\documentclass[runningheads,a4paper]{llncs}
\usepackage[utf8]{inputenc}
\usepackage[T1]{fontenc}
\usepackage{csquotes}
\usepackage[english]{babel}
\usepackage[backend=bibtex,natbib,hyperref=true,maxnames=2,style=authoryear-comp]{biblatex}
\usepackage{amsmath}
\usepackage{booktabs}
\usepackage{tabularx}
\newcommand{\ra}[1]{\renewcommand{\arraystretch}{#1}}
\usepackage{enumitem}
\usepackage{graphicx}
\usepackage{eurosym}
\usepackage{amssymb}
\usepackage{adjustbox}

\usepackage[hidelinks]{hyperref}
\usepackage[capitalise]{cleveref}
\usepackage{subfigure}
\begin{document}
\frontmatter     
\pagestyle{headings} 
\mainmatter       
\title{Semantic video segmentation \\ for autonomous driving}
\titlerunning{Semantic video segmentation for autonomous driving}
\author{Chau, Minh Triet}
\authorrunning{Chau, Minh Triet}  
\institute{Universit\"at Bonn\\
}

\maketitle       

\begin{abstract}
  We aim to solve semantic video segmentation in autonomous driving, namely road detection in real time video, using techniques discussed in \citep{DBLP:journals/corr/ShelhamerRHD16}. While fully convolutional network gives good result, we show that the speed can be halved while preserving the accuracy. The test dataset being used is KITTI, which consists of real footage from Germany's streets \citep{Geiger2013IJRR}.
\end{abstract}

\section{Introduction}
Full convolution network has been successfully applied for recognition tasks. However, in the field of video semantic segmentation, it would make little sense to do the inference in every frame. Rather one should exploit the continuity of features through frame. Even in the in the field of image semantic segmentation, semantic segmentation has some difficulties with a pure convolution approach. This lab aims to take the best of both world from \citep{DBLP:journals/corr/ShelhamerRHD16} for Clockwork convolution network and \citep{DBLP:journals/corr/CaesarUF16} for Region-based semantic segmentation, then test on the KITTI dataset.
dataset \citep{Geiger2013IJRR}

\section{Related work}
\subsection{Fully Convolutional Networks (FCN)}
 A fully convolutional network (FCN) is a model designed for pixelwise prediction \citep{DBLP:journals/corr/ShelhamerLD16}. Every layer in an FCN computes a local operation, such as convolution or pooling, on relative spatial coordinates. This locality makes the network capable of handling inputs of any size while producing output of corresponding dimensions. These networks are fast at inference and learning time.
 
 \subsection{Spatiotemporal filtering}
 For video classification, networks can be integrated over time by fusion of frame features \citep{KarpathyCVPR14}. Recurrence can capture long-term dynamics and propagate state across time, as in the popular long short-term memory (LSTM) \citep{Hochreiter:1997:LSM:1246443.1246450}. Joint convolutional-recurrent networks filter within frames and recur across frames: the long-term recurrent convolutional network [17] fuses frame features by LSTM for activity recognition and captioning. 
 
 \subsection{Network Acceleration} Video demands faster segmentation speed. The spatially dense operation of the FCN amortizes the computation of overlapping receptive fields common to contemporary architectures. However, the standard FCN does nothing to temporally amortize the computation of sequential inputs. The clockwork presented in \citep{DBLP:journals/corr/ShelhamerRHD16} holds for all layered architectures whatever the speed/quality operating point chosen.

\section{Clockwork convolution network}
\subsection{Structure}
In the domain of spatiotemporal segmentation, recurrent neuron network has been proved to be useful. However, it has problem with exploding and vanishing gradient through long sequence of frames. Thus, the appearance of Clockwork network to solve that problem. 
\begin{figure}[!htbp]
  \centering
  \includegraphics[width=0.75\linewidth]{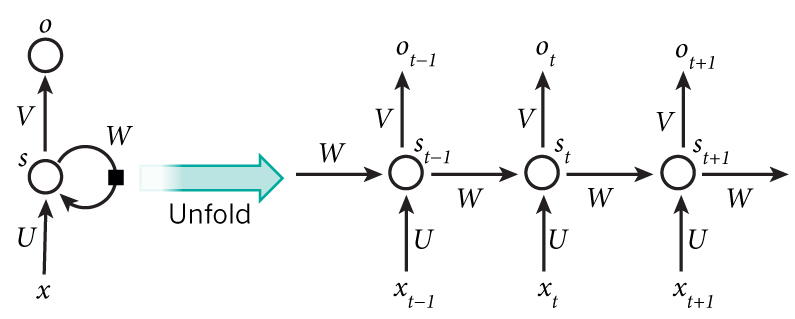}  
  \caption{A recurrent neural network and the unfolding in time of the computation involved in its forward computation. Source: Nature}
  \label{fig::rnn}
\end{figure}
\begin{figure}[!htbp]
  \centering
  \includegraphics[width=0.75\linewidth]{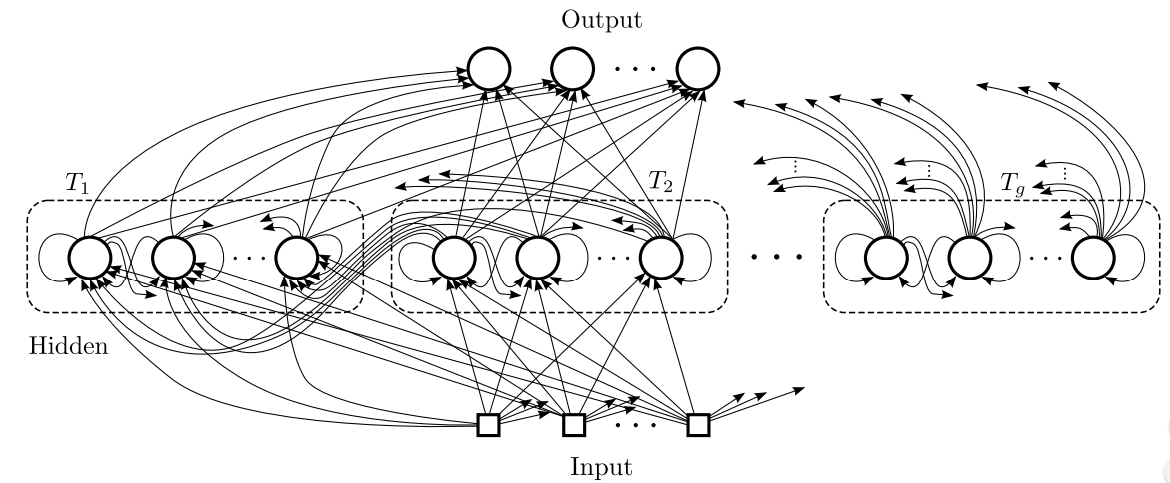}  
  \caption{The model of clockwork RNN. The hidden layer is partitioned into \(g\) modules each with its own clock rate. Within each module the neurons are fully interconnected. Neurons in faster module i are connected to neurons in a slower module j only if a clock period \(T_i < T_j\) \citep{DBLP:journals/corr/KoutnikGGS14}}
\end{figure}
\subsection{Clockwork Recurrent Neuron Network}
Noted that the design of RNN, neurons in faster module i are connected to neurons in a slower module j only if a clock period \(T_i < T_j\). That intuition also fits with the fact in CNN that the semantic of deep layers is relevant across frames whereas the shallow layers varies. Persisting these deep features over time can be seen as a temporal skip connection. That is why man can implement a skip mechanism, known as clockwork, to route the processing progress. A clock is associated with a fixed or adaptable interval. It fires when the compute progress is allowed to pass through its layers that correspond 
\begin{figure}[htbp]
    \centering
    \includegraphics[width=0.75\linewidth]{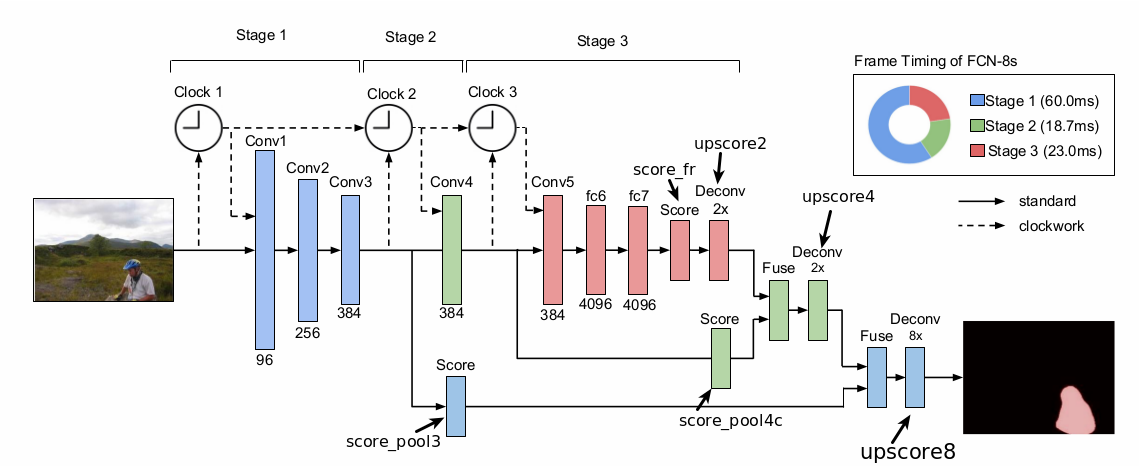}
    \caption{The annotated network with same variable name from the code}
    \label{fig:my_label}
\end{figure}

To determine how much time should a clock fire, the execution of a stage on a given frame is determined by either a fixed clock rate or adaptive (data-driven). In the following examples, we further clarify the idea using adaptive clockwork, which we are convinced to be better (confirm section~\ref{ss:adaptive}).

\begin{figure*}[!htbp]
    \centering
    \subfigure[]{
    	\includegraphics[width=0.75\linewidth]{pic/frames.png}
        \label{fig:stage0}
    }
    \subfigure[]{
 	   \includegraphics[width=0.75\linewidth]{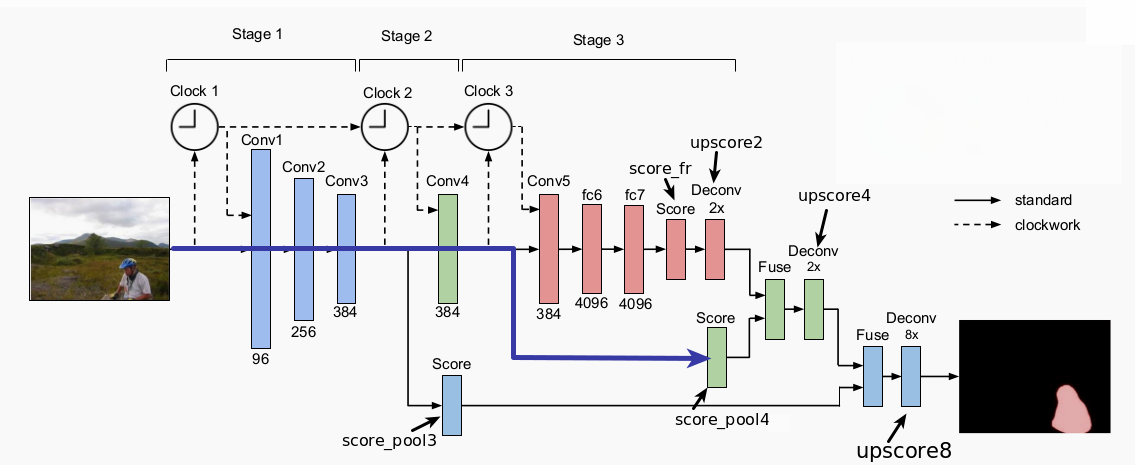}
    	\label{fig:stage1}
    }
    \subfigure[]{
    	\includegraphics[width=0.75\linewidth]{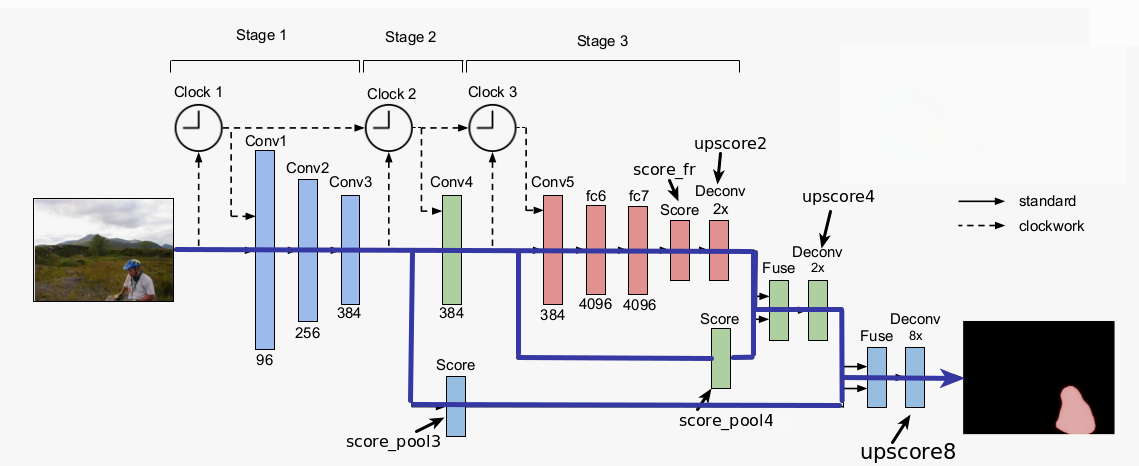}
    	\label{fig:stage2}
    }    
    \caption{~\ref{fig:stage0} Frame configuration.~\ref{fig:stage1} At the beginning of the video, the network do a full segmentation, and save the score of layer $score\_pool4$ to variable $prev\_score$.~\ref{fig:stage2} Since the difference between frame 1 and frame 2 are low, the difference between $score\_pool4$ at frame 2 and $prev\_score$ is low. The network stop at layer $score\_pool4$. $prev\_score$ stays the same.}
\end{figure*}

\subsection{Clockwork}
\subsubsection{Fixed rate}
These clock rates are free parameters in the schedule for exchanging inference speed and accuracy. We again divide the network into three stages, and compare rates for the stages. The exponential clockwork schedule is the natural choice of halving the rate at each stage for more efficiency.
\subsubsection{Adaptive clockwork} \label{ss:adaptive}
After some experiences, we are convinced that adaptive is more suitable for segmenting videos collected in driving because there is no telling if the driver suddenly speed up or slow down in a short time interval. A fixed rate clockwork cannot capture this unexpected nature of behavior on the road, while adaptive clocks fire based on the input and network state, resulting in a responsive schedule that varies with the dynamism of the scene.

\section{Result}
\subsection{KITTI Dataset}
The KITTI dataset has been recorded from a moving platform while driving in and around Karlsruhe, Germany. It consists of three different types of roads UU, UM, UMM \ref{table:kitti}. Two main challenges of KITTI dataset beside the lacking of ground truth annotation are (1) lane detection, which requires our pre-trained network to unlearn the concept of road and learn  the concept of the current lane of the car instead, and (2) The road being interested in, which appears in UMM and UM road, annotated in magenta and the other roads annotated in black.

\begin{table}[htbp]
        \centering
        \caption{Size comparison between KITTI and other datasets}
        \ra{1.3}
        \begin{tabular}{@{}ll@{}}
            \toprule
            Dataset & Content \\
            \midrule
            Pascal & 736 image subset of the PASCAL VOC 2011 \\
            Youtube & 10,167 frames from 126 shots \\
            NYU Dataset & 1449 densely labeled indoor images \\
            KITTI Road detection & 289 training and 290 test images\\
            \bottomrule
        \end{tabular}
\end{table}

\begin{table}[htbp]
    \ra{1.3}
    \centering 
    \label{table:kitti}
    \caption{Clarification of the KITTI dataset}
    \begin{tabular}{@{}l c  c  c  c  c @{}}
    \toprule
    Name & Description & Training & Testing & Total & Frame and ground truth \\
    \midrule
    UU & One way road & 68 & 30 & 98 & \includegraphics[width=0.25\linewidth]{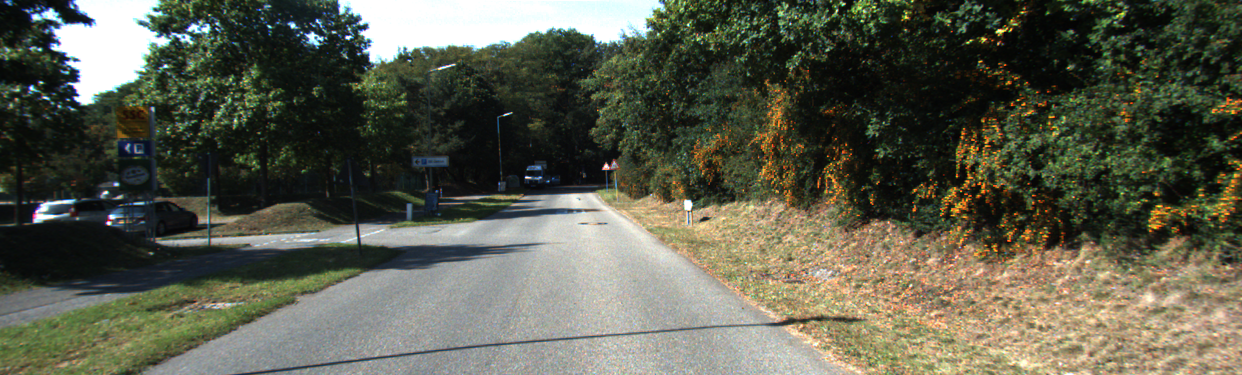}    
    \includegraphics[width=0.25\linewidth]{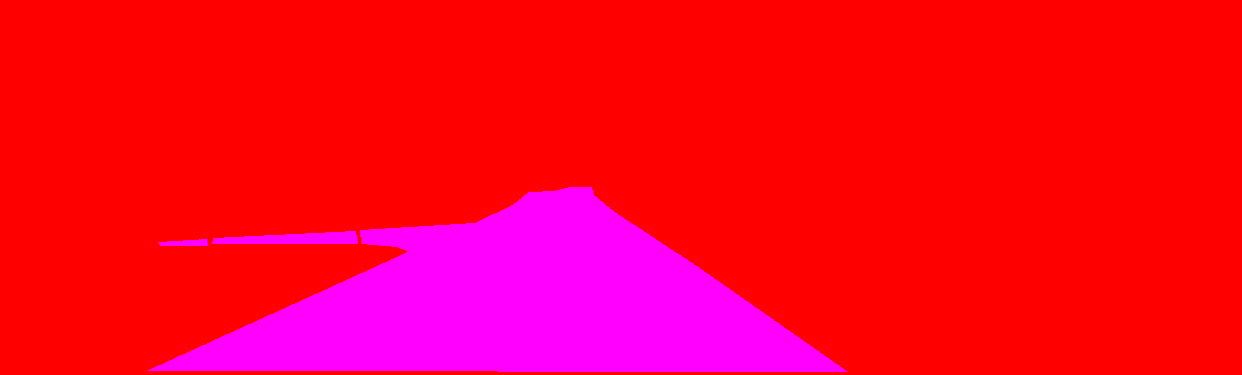} \\
    UM & Two-way road & 65 & 30 & 95 & \includegraphics[width=0.25\linewidth]{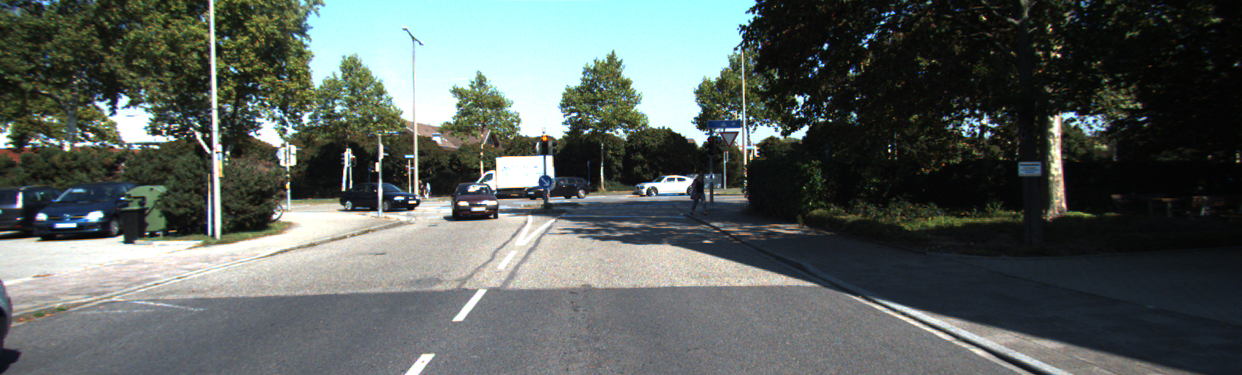} \includegraphics[width=0.25\linewidth]{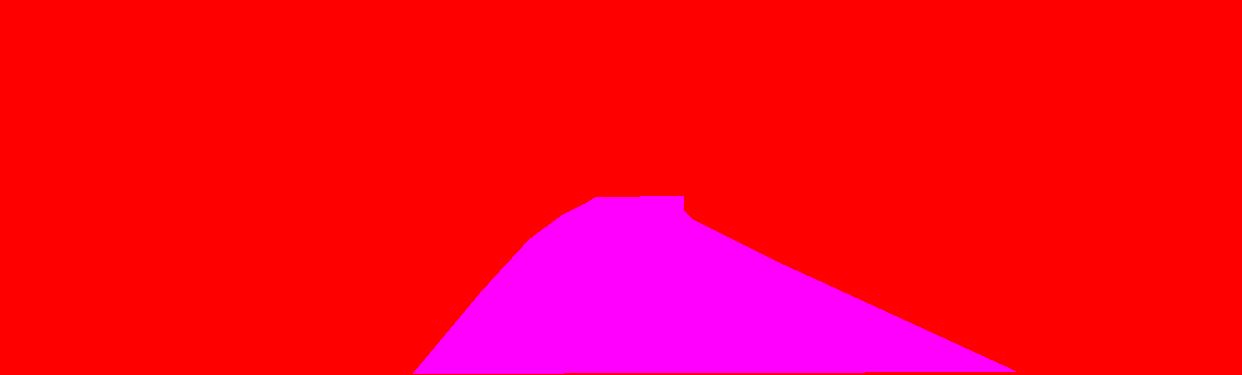} \\
    UMM & Multilane road & 66 & 30 & 96 & \includegraphics[width=0.25\linewidth]{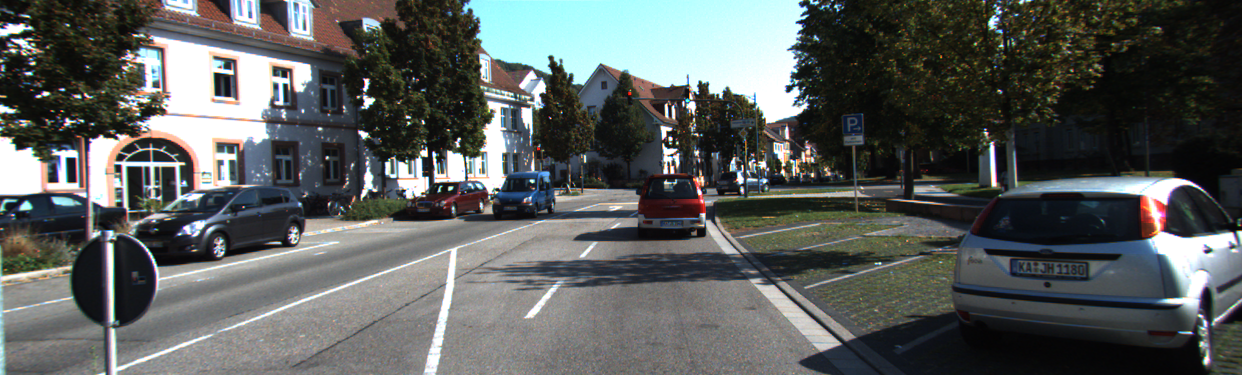} \includegraphics[width=0.25\linewidth]{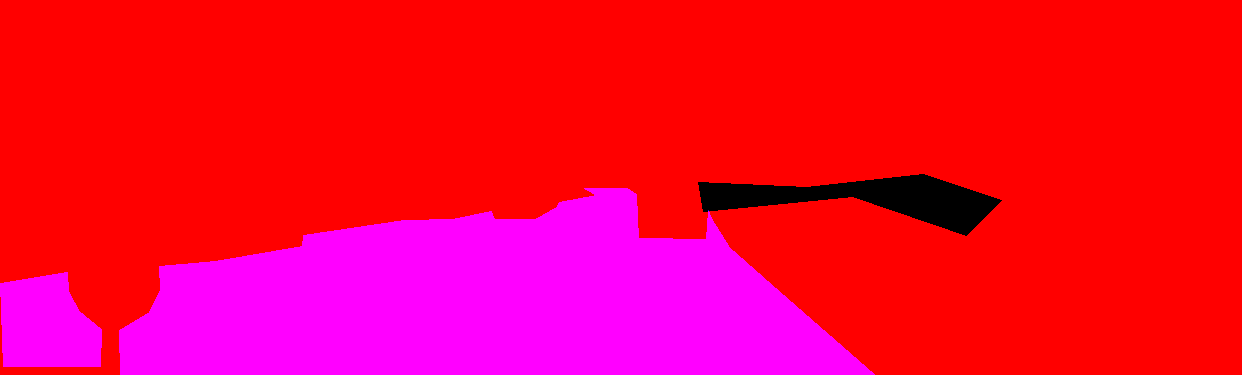}\\
    \bottomrule
    \end{tabular}    
\end{table}

\subsection{Test result}
The network used \texttt{fcn8} structure, which was proposed in \citep{DBLP:journals/corr/ShelhamerRHD16} and pretrained weight on Cityscape dataset. It has been further fine tuned with Kitti dataset with the following test result.    

\begin{table}[htbp]
        \centering
        \ra{1.3}
        \caption{Test result using KITTI dataset}
        \begin{tabular}{@{} lccc @{}}
        \toprule
        \texttt{acc} (\%)&		 \texttt{cl\_acc} (\%)&		 \texttt{mIU} (\%)& \texttt{fwIU} (\%)\\
        \midrule
        92.53 &	 87.31&	 21.92 & 87.08 \\
        \bottomrule
        \end{tabular}
        \begin{align*}
        	\texttt{acc} &: \textnormal{Total accuracy}\\
            \texttt{cl acc} &: \textnormal{Mean accuracy for each classes (Road and non road)}\\
            \texttt{mIU} &: \textnormal{ Intersection over Union = true positive / (true positive + false positive + false negative)} \\
            \texttt{fwIU} &: \textnormal{Frequency weighted IU} = \frac{\textnormal{pixels of a class}}{\textnormal{total pixels}}\texttt{mIU}
        \end{align*} 
    \end{table}

We also compare with other methods in Table \ref{tab:my_label}.

\begin{table}[!htpb]
\caption{Semantic segmentation results on UMM Road and UU Road}
    \label{tab:my_label}
    \ra{1.3}
\adjustbox{max width=\textwidth}{%
    \begin{tabular}{@{}lrrrrrr@{}}
    \toprule
    Method & F1-score & Avr Precision (\%) & Precision (\%) & Recall (\%) & False Positive (\%) & False Negative (\%) \\
    \midrule
    \textbf{UMMRoad} &&&&&& \\
    Baseline & 85.32  & 75.67 & 93.47 & 78.47  & 1.77 & 21.53 \\
    Clockwork Network & 91.10  & 91.81 & 88.94 & 93.36 & 3.75 & 6.64 \\
    Full Convolution & 92.78 & 91.90 & 91.85 & 94.05 & 2.41 & 5.95 \\
    \textbf{UURoad} &&&&&& \\
    Baseline & 74.84 & 60.85 & 75.74 & 73.97 & 3.69 & 26.03 \\        Clockwork Network & 81.28 & 85.23 & 80.14 & 82.45 & 3.18 & 17.55 \\
        Full Convolution & 88.94 & 86.22 & 81.25 & 84.25 & 2.55 & 15.75 \\
    \bottomrule
    \end{tabular}}
\end{table}

The network fully annotated in 24.392047 seconds. Annotating every frames without clockwork network takes 55.790964 seconds.

\section{Conclusion}
The proposed method in \citep{DBLP:journals/corr/ShelhamerRHD16} is is an extension of fully convolutional networks for image semantic segmentation to video semantic segmentation using the stability of deep level features in videos. Thanks to the idea of Recurrent Neural Nets, the computation can be stopped at earlier layers, and be preserved to be fused with results in the previous frames. Thus, the speed has been roughly halved while the metrics like precision, recall, accuracy are similar to the full frame segmentation. The result is better in videos that took place in one way road, due to the fact of the  fuzziness of road being interested in in marked multilane roads.
\printbibliography
\end{document}